\documentclass[sigconf,nonacm]{acmart}
  
\usepackage{mathtools}
\usepackage{amssymb}
\usepackage{graphicx}
\usepackage[english]{babel}
\usepackage{float}
\usepackage{listings}
\usepackage[most]{tcolorbox}
\usepackage{pifont}
\usepackage{booktabs}
\usepackage{array}
\usepackage{minted}
\usepackage{enumitem}
\usepackage{tikz}
\usepackage{multirow}

\newcommand{\rotcell}[1]{%
  \raisebox{0pt}[0pt][0pt]{\rotatebox{90}{#1}}%
}

\newcommand{\benchName}{\textbf{GeoBenchLLM}}

\AtBeginDocument{%
  }

\setcopyright{acmlicensed}
\copyrightyear{2026}
\acmYear{2026}
\acmDOI{XXXXXXX.XXXXXXX}

\begin{document}

\title{\benchName: A Comprehensive Benchmark for Evaluating LLMs on Geo-Related Tasks}

\author{Rodrigo Ferreira Rodrigues}
\email{rodrigo.ferreira-rodrigues@utoulouse.fr }
\affiliation{%
  \institution{University of Toulouse}
  \city{Toulouse}
  \country{France}
}

\author{Karim Radouane}
\email{karim.radouane@irit.fr }
\affiliation{%
  \institution{University of Toulouse}
  \city{Toulouse}
  \country{France}
}

\author{Jose G Moreno}
\email{jose.moreno@irit.fr }
\affiliation{%
  \institution{University of Toulouse}
  \city{Toulouse}
  \country{France}
}

\author{Lynda Tamine}
\email{lynda.tamine@irit.fr }
\affiliation{%
  \institution{University of Toulouse}
  \city{Toulouse}
  \country{France}
}

\begin{abstract}
In the context of geodata, existing Large Language Models have often been studied 
in a homogeneous setting, which has considerably limited insights into their 
generalization capabilities. In this paper, we present \benchName, a comprehensive 
benchmark for probing LLMs on geo-related tasks. We leverage a careful selection 
of twelve publicly available datasets from diverse geo-related tasks and domains, 
and evaluate a set of LLMs on geo-spatial and temporal understanding using our 
benchmark. Our results show that reasoning and size have a strong impact on overall 
performance. \benchName\ is publicly available at \url{https://github.com/Rfr2003/GeoBenchLLM}. 
\end{abstract}

\keywords{Benchmark, Large Language Models}

\maketitle
\fancyhead{}

\section{Introduction}

Geo-related tasks refer to problems involving geographic entities or concepts, usually requiring spatial operations to be solved. These tasks range from factoid questions (e.g., \textit{Is Paris north of Toulouse?}) to pathfinding problems (e.g., \textit{Show me a route from Paris to Toulouse}), and can be very challenging for QA systems to solve. Mai et al. \cite{gqachallenges} attribute these challenges to geometric uncertainty as well as language variability and vagueness.

Large Language Models (LLMs) have garnered significant attention due to their extensive capabilities in processing language and handling complex tasks across diverse domains. In the geographical field, researchers were among the first to assess LLMs' abilities in geo-related tasks. Manvi et al. \cite{regpop} evaluated LLMs' capacities in geographical regression tasks, while Li et al. \cite{mapqa} proposed a dataset to evaluate LLMs on a place prediction task involving Points of Interest (POIs).

An effort from various researchers has already been made to build benchmarks (Table \ref{tab:comparaison}) in order to evaluate LLMs across a wide range of geo-related tasks, but they are either too small (Xu et al. \cite{evaluatingllm}), do not cover enough major tasks (STBench \cite{stbench}), or are limited to a specific context (CityEval \cite{citygpt}). To address the lack of a more comprehensive geographical evaluation benchmark, we introduce \benchName, a highly accessible and comprehensive benchmark for assessing the intrinsic knowledge of LLMs on geo-related tasks. Our benchmark brings together twelve datasets, further divided and transformed into seventeen purely textual subdatasets (e.g., a subset of a dataset containing only questions related to a single task; this subset can be identical to the dataset), covering eight different tasks. We adopt the higher-level classification used by Xu et al. \cite{evaluatingllm} in their geographic benchmark to categorize the tasks into three main cognitive levels: \textit{Knowledge}, \textit{Reasoning}, and \textit{Application}.

We evaluate our benchmark using models from the Qwen family, with sizes ranging from 0.6B to 8B, both with and without thinking enabled, as well as larger models, namely GPT-OSS-20B and 120B, with limited thinking capabilities. To evaluate these subdatasets, we also introduce a set of manually designed metrics. Figure \ref{fig:code} shows that LLMs up to 120B parameters are able to succeed on the benchmark, but smaller models may close the gap only if thinking is enabled, especially for tasks belonging to the Reasoning and Application cognitive levels.

\section{Related works}

This section presents existing benchmarks and datasets used for assessing LLMs abilities on geo-related tasks.

\begin{table}[]
\caption{Comparison of our \benchName\ with several benchmarks and datasets on geo-related domains. Knowledge (Know.), Reasoning (Reas.), Application (Appl.), Generative (Gen.), Regression (Reg.), Yes/No (Y/N).}
\vspace{-0.3cm}
\resizebox{0.48\textwidth}{!}{%
\begin{tabular}{ll|c|c|c|c|c|r}
\hline
\multicolumn{2}{l|}{\textbf{Tasks ($\rightarrow$)}}                            & \textbf{Know.} & \textbf{Reas.} & \textbf{Appl.} & \textbf{Scale} & \textbf{Format}             & \multicolumn{1}{l}{\textbf{\# Examples}} \\ \hline
\multicolumn{1}{l|}{\textbf{Benchmarks}} & GeoBenchmark \cite{geobenchmark}           & \ding{51}      & \ding{51}      & \ding{56}      & Country           & Y/N, MCQ                         & 39 378 \\
\multicolumn{1}{l|}{\textbf{}}  & CityEval \cite{citygpt}           & \ding{51}      & \ding{51}      & \ding{51}      & City           & MCQ                         & 955 463                                  \\
\multicolumn{1}{l|}{\textbf{}}           & Xu et al. \cite{evaluatingllm}    & \ding{51}      & \ding{51}      & \ding{51}      & World          & Gen., MCQ, Reg., Y/N & 900                                      \\
\multicolumn{1}{l|}{\textbf{}}           & STBench \cite{stbench}            & \ding{51}      & \ding{51}      & \ding{51}      & World          & MCQ                         & 82 750                                   \\ \hline
\multicolumn{1}{l|}{\textbf{Datasets}}   & bAbI (tasks 17 \& 19) \cite{babi} & \ding{56}      & \ding{51}      & \ding{56}      & -              & Gen.                 & 4 000                                    \\
\multicolumn{1}{l|}{\textbf{}}           & MapQA \cite{mapqa}                & \ding{51}      & \ding{56}      & \ding{56}      & State          & Gen., Reg.      & 3 154                                    \\ \hline
\multicolumn{1}{l|}{\textbf{Ours}}       & \benchName                        & \ding{51}      & \ding{51}      & \ding{51}      & World          & Gen., MCQ, Reg., Y/N & 421 041                                  \\ \hline
\end{tabular}%
}
\label{tab:comparaison}

\vspace{-.5cm}
\end{table}

\subsection{Benchmarks}

Geo-Benchmarks offer a setting to assess LLMs in a range of geo-related tasks, often providing metrics and/or baselines in addition to evaluation datasets.

GeoBenchmark by Abayomi-Alli et al. \cite{geobenchmark} aims to evaluate geographic commonsense along three core spatial concepts: direction, distance, and topology, using data extracted from YAGO2geo\footnote{\url{https://yago2geo.di.uoa.gr/}} and Ordnance Survey ward geometries. Despite providing an in-depth assessment of LLMs' geospatial capabilities at the \textit{Knowledge} and \textit{Reasoning} levels in a Yes/No and MCQ format, it doesn't cover tasks from the \textit{Application} cognitive level.

STBench by Li et al. \cite{stbench} contains around 80,000 author-generated questions derived mainly from the Yelp dataset\footnote{\url{https://www.yelp.com/dataset}}, directed towards evaluating the temporal characteristics of geographic questions. However, it heavily focuses on spatial reasoning tasks and lacks variety, missing \textit{Coordinates Prediction}, \textit{Regression}, and \textit{Complex Scenario QA}. Xu et al. \cite{evaluatingllm} provides a more diverse benchmark, but focuses on evaluating LLMs' ability to use external tools (e.g., code, APIs) rather than their intrinsic knowledge. Although the 900 questions, collected from public datasets, Wikipedia, and geography textbooks, are of high quality, their number is very limited, and the benchmark still lacks critical tasks like \textit{Coordinates Prediction}. CityEval by Feng et al. \cite{citygpt}, part of the CityGPT framework, focuses on enhancing LLMs' understanding of urban space using an instruction-tuning dataset \textit{CityInstruction}. It proposes a large number of questions covering most of our tasks, with the exception of \textit{Regression} and \textit{Complex Scenario QA}, but remains limited to an urban context. Finally,\cite{geobenchmark}, \cite{stbench} and \cite{citygpt} only propose questions in a Yes/No or MCQ format, which, while easier to evaluate, restricts the model to a rare-case scenario where choices are limited to a small set of answers.

\subsection{Datasets}

bAbI \cite{babi} was introduced by Facebook to evaluate language models on their ability to handle question answering tasks. It covers a wide range of tasks going from \textit{Single Supporting Fact} to \textit{Simple Negation and Indefinite Knowledge}. Only two tasks can be labelled as geographic, these are the tasks 17 and 19, focusing on positional reasoning and pathfinding respectively.

Li et al. \cite{mapqa} proposed MapQA, a dataset focused on answering geospatial questions between POIs located in Southern California and Illinois. The data is retrieved by the authors from OpenStreetMap. The questions focus on predicting a place name knowing its amenity type and its spatial relationship in regards to another place. It also provides regression questions asking the distance between two POIs.

\subsection{Motivation for GeoBenchLLM}

Table \ref{tab:comparaison} summarizes the differences between our benchmark and others. The datasets don't cover of course all cognitive levels, as their goal is limited to one or two tasks. As for the benchmarks, we can see that they possess at least one task belonging to each of the three cognitive levels, with the exception of GeoBenchmark, only targeting tasks of the \textit{Knowledge} and \textit{Reasoning} levels. However, their scale is either too narrow (\textit{CityGPT}), or their format is limited to MCQ (\textit{CityGPT} and \textit{STBench}), restricting the LLM's capacities. Even when the benchmark seems to check all the boxes (\textit{Xu et al.}), its size is too small to really conduct a comprehensive assessment of the geographic capabilities of LLMs.

To fill this gap, we propose a complete benchmark able to probe LLMs on geo-related tasks across three cognitive levels. Our benchmark contains 421,041 questions at a world scale, in all formats: Generative, Regression, Yes/No questions, and MCQ. To be able to assess fairly and accurately all these questions in different formats, we also introduce new metrics detailed in Section \ref{metrics}.

\section{\benchName}
In this section, we present our benchmark made from twelve public datasets. We explain the steps of data collection and then describe each dataset along with the tasks they belong to. Further details about the benchmark are given in Table \ref{tab:description}.

\begin{table*}[]
\resizebox{\textwidth}{!}{%
\begin{tabular}{lllcccrrr|cccccccccccccccccccc}
         &                                             &                                                   & \multicolumn{1}{l}{}                     & \multicolumn{1}{l}{}                    & \multicolumn{1}{l}{}                   & \multicolumn{1}{l}{}         &        & \multicolumn{1}{l|}{}                 & \multicolumn{3}{c|}{\textbf{Avg Word}}        & \multicolumn{4}{c|}{\textbf{Accuracy}}                                                                 &                                         &                                         &                                      &                                          &                                   &                                      &                                          &                                  &                                  &                                  &                                  &                                        &                                               \\ 
\multicolumn{1}{l|}{\textbf{Cogn. lvl}}                     & \multicolumn{1}{l|}{\textbf{Tasks}}         & \multicolumn{1}{l|}{\textbf{SubDatasets}}            & \multicolumn{1}{l|}{\rotcell{\textbf{Syn.}}} & \multicolumn{1}{c|}{\rotcell{\textbf{Sce.}}} & \multicolumn{1}{c|}{\rotcell{\textbf{Trs.}}} & \multicolumn{1}{c|}{\textbf{\# Train}} & \multicolumn{1}{c|}{\textbf{\# Dev}} & \multicolumn{1}{c|}{\textbf{\# Test}} &  \multicolumn{1}{c|}{Sce}    &  \multicolumn{1}{c|}{Ques}   &  \multicolumn{1}{c|}{Ans}  & \multicolumn{1}{c|}{@1} & \multicolumn{1}{c|}{@3} & \multicolumn{1}{c|}{@5} & \multicolumn{1}{c|}{@10} & \multicolumn{1}{c|}{\rotcell{\textbf{C Acc}}} & \multicolumn{1}{c|}{\rotcell{\textbf{Prec}}} & \multicolumn{1}{c|}{\rotcell{\textbf{Rec}}} & \multicolumn{1}{c|}{\rotcell{\textbf{P-R Mean}}} & \multicolumn{1}{c|}{\rotcell{\textbf{Median}}} & \multicolumn{1}{c|}{\rotcell{\textbf{Bleu-1}}} & \multicolumn{1}{c|}{\rotcell{\textbf{B-Sco}}} & \multicolumn{1}{c|}{\rotcell{\textbf{C Rat}}} & \multicolumn{1}{c|}{\rotcell{\textbf{S Rat}}} & \multicolumn{1}{c|}{\rotcell{\textbf{O Rat}}} & \multicolumn{1}{c|}{\rotcell{\textbf{F Rat}}} & \multicolumn{1}{c|}{\rotcell{\textbf{Dist}}} & \multicolumn{1}{c}{\rotcell{\textbf{U Acc}}} \\ \midrule

\multicolumn{1}{l|}{\textbf{Knowledge}}   & \multicolumn{1}{l|}{Coordinates Prediction} & \multicolumn{1}{l|}{GeoQuestions1089\_coord \cite{geoquestions}}      & \multicolumn{1}{c|}{\ding{56}}                   & \multicolumn{1}{c|}{\ding{56}}    & \multicolumn{1}{c|}{\ding{51}}               & \multicolumn{1}{r|}{-}                 & \multicolumn{1}{r|}{-}               & 87                                    & -      & 5.27   & \multicolumn{1}{c|}{2.00}   & \multicolumn{1}{c|}{\ding{56}}  & \multicolumn{1}{c|}{\ding{56}}  & \multicolumn{1}{c|}{\ding{56}}  & \multicolumn{1}{c|}{\ding{56}}   & \multicolumn{1}{c|}{\ding{73}}                  & \multicolumn{1}{c|}{\ding{56}}                  & \multicolumn{1}{c|}{\ding{56}}               & \multicolumn{1}{c|}{\ding{56}}                   & \multicolumn{1}{c|}{\ding{56}}            & \multicolumn{1}{c|}{\ding{56}}               & \multicolumn{1}{c|}{\ding{56}}                   & \multicolumn{1}{c|}{\ding{56}}           & \multicolumn{1}{c|}{\ding{56}}           & \multicolumn{1}{c|}{\ding{56}}           & \multicolumn{1}{c|}{\ding{56}}           & \multicolumn{1}{c|}{\ding{56}}                 & \multicolumn{1}{c}{\ding{56}}                        \\ \cline{2-29} 
\multicolumn{1}{l|}{}                     & \multicolumn{1}{l|}{Yes/No questions}       & \multicolumn{1}{l|}{GeoQuestions1089\_YN \cite{geoquestions}}         & \multicolumn{1}{c|}{\ding{56}}                   & \multicolumn{1}{c|}{\ding{56}}             & \multicolumn{1}{c|}{\ding{51}}       & \multicolumn{1}{r|}{-}                 & \multicolumn{1}{r|}{-}               & 181                                   & -      & 7.19   & \multicolumn{1}{c|}{1.00}   & \multicolumn{1}{c|}{\ding{73}}  & \multicolumn{1}{c|}{\ding{56}}  & \multicolumn{1}{c|}{\ding{56}}  & \multicolumn{1}{c|}{\ding{56}}   & \multicolumn{1}{c|}{\ding{56}}                  & \multicolumn{1}{c|}{\ding{56}}                  & \multicolumn{1}{c|}{\ding{56}}               & \multicolumn{1}{c|}{\ding{56}}                   & \multicolumn{1}{c|}{\ding{56}}            & \multicolumn{1}{c|}{\ding{56}}               & \multicolumn{1}{c|}{\ding{56}}                   & \multicolumn{1}{c|}{\ding{56}}           & \multicolumn{1}{c|}{\ding{56}}           & \multicolumn{1}{c|}{\ding{56}}           & \multicolumn{1}{c|}{\ding{56}}           & \multicolumn{1}{c|}{\ding{56}}                 & \multicolumn{1}{c}{\ding{56}}                        \\ \cline{2-29} 
\multicolumn{1}{l|}{}                     & \multicolumn{1}{l|}{Regression}             & \multicolumn{1}{l|}{GeoQuestions1089\_regression \cite{geoquestions}} & \multicolumn{1}{c|}{\ding{56}}                   & \multicolumn{1}{c|}{\ding{56}}       & \multicolumn{1}{c|}{\ding{51}}             & \multicolumn{1}{r|}{-}                 & \multicolumn{1}{r|}{-}               & 231                                   & -      & 8.16   & \multicolumn{1}{c|}{1.27}   & \multicolumn{1}{c|}{\ding{56}}  & \multicolumn{1}{c|}{\ding{56}}  & \multicolumn{1}{c|}{\ding{56}}  & \multicolumn{1}{c|}{\ding{56}}   & \multicolumn{1}{c|}{\ding{56}}                  & \multicolumn{1}{c|}{\ding{51}}                  & \multicolumn{1}{c|}{\ding{51}}               & \multicolumn{1}{c|}{\ding{51}}                   & \multicolumn{1}{c|}{\ding{73}}            & \multicolumn{1}{c|}{\ding{56}}               & \multicolumn{1}{c|}{\ding{56}}                   & \multicolumn{1}{c|}{\ding{56}}           & \multicolumn{1}{c|}{\ding{56}}           & \multicolumn{1}{c|}{\ding{56}}           & \multicolumn{1}{c|}{\ding{56}}           & \multicolumn{1}{c|}{\ding{56}}                 & \multicolumn{1}{c}{\ding{56}}                        \\
\multicolumn{1}{l|}{}                     & \multicolumn{1}{l|}{}                       & \multicolumn{1}{l|}{GeoQuery\_regression \cite{geoquery} \cite{texttosql}}         & \multicolumn{1}{c|}{\ding{56}}                   & \multicolumn{1}{c|}{\ding{56}}          & \multicolumn{1}{c|}{\ding{51}}          & \multicolumn{1}{r|}{182}               & \multicolumn{1}{r|}{17}              & 89                                    & -      & 8.77   & \multicolumn{1}{c|}{1.87}   & \multicolumn{1}{c|}{\ding{56}}  & \multicolumn{1}{c|}{\ding{56}}  & \multicolumn{1}{c|}{\ding{56}}  & \multicolumn{1}{c|}{\ding{56}}   & \multicolumn{1}{c|}{\ding{56}}                  & \multicolumn{1}{c|}{\ding{51}}                  & \multicolumn{1}{c|}{\ding{51}}               & \multicolumn{1}{c|}{\ding{51}}                   & \multicolumn{1}{c|}{\ding{73}}            & \multicolumn{1}{c|}{\ding{56}}               & \multicolumn{1}{c|}{\ding{56}}                   & \multicolumn{1}{c|}{\ding{56}}           & \multicolumn{1}{c|}{\ding{56}}           & \multicolumn{1}{c|}{\ding{56}}           & \multicolumn{1}{c|}{\ding{56}}           & \multicolumn{1}{c|}{\ding{56}}                 & \multicolumn{1}{c}{\ding{56}}                        \\ \cline{2-29} 
\multicolumn{1}{l|}{}                     & \multicolumn{1}{l|}{Place prediction}       & \multicolumn{1}{l|}{GeoQuestions1089\_place \cite{geoquestions}}      & \multicolumn{1}{c|}{\ding{56}}                   & \multicolumn{1}{c|}{\ding{56}}           & \multicolumn{1}{c|}{\ding{51}}         & \multicolumn{1}{r|}{-}                 & \multicolumn{1}{r|}{-}               & 455                                   & -      & 8.53   & \multicolumn{1}{c|}{701.32} & \multicolumn{1}{c|}{\ding{56}}  & \multicolumn{1}{c|}{\ding{56}}  & \multicolumn{1}{c|}{\ding{56}}  & \multicolumn{1}{c|}{\ding{56}}   & \multicolumn{1}{c|}{\ding{56}}                  & \multicolumn{1}{c|}{\ding{51}}                  & \multicolumn{1}{c|}{\ding{51}}               & \multicolumn{1}{c|}{\ding{51}}                   & \multicolumn{1}{c|}{\ding{73}}             & \multicolumn{1}{c|}{\ding{51}}               & \multicolumn{1}{c|}{\ding{51}}                   & \multicolumn{1}{c|}{\ding{56}}           & \multicolumn{1}{c|}{\ding{56}}           & \multicolumn{1}{c|}{\ding{56}}           & \multicolumn{1}{c|}{\ding{56}}           & \multicolumn{1}{c|}{\ding{56}}                 & \multicolumn{1}{c}{\ding{56}}                        \\
\multicolumn{1}{l|}{}                     & \multicolumn{1}{l|}{}                       & \multicolumn{1}{l|}{GeoQuery\_place \cite{geoquery} \cite{texttosql}}              & \multicolumn{1}{c|}{\ding{56}}                   & \multicolumn{1}{c|}{\ding{56}}        & \multicolumn{1}{c|}{\ding{51}}            & \multicolumn{1}{r|}{346}               & \multicolumn{1}{r|}{33}              & 184                                   & -      & 8.47   & \multicolumn{1}{c|}{10.65}  & \multicolumn{1}{c|}{\ding{56}}  & \multicolumn{1}{c|}{\ding{56}}  & \multicolumn{1}{c|}{\ding{56}}  & \multicolumn{1}{c|}{\ding{56}}   & \multicolumn{1}{c|}{\ding{56}}                  & \multicolumn{1}{c|}{\ding{51}}                  & \multicolumn{1}{c|}{\ding{51}}               & \multicolumn{1}{c|}{\ding{51}}                   & \multicolumn{1}{c|}{\ding{73}}            & \multicolumn{1}{c|}{\ding{51}}               & \multicolumn{1}{c|}{\ding{51}}                   & \multicolumn{1}{c|}{\ding{56}}           & \multicolumn{1}{c|}{\ding{56}}           & \multicolumn{1}{c|}{\ding{56}}           & \multicolumn{1}{c|}{\ding{56}}           & \multicolumn{1}{c|}{\ding{56}}                 & \multicolumn{1}{c}{\ding{56}}                        \\
\multicolumn{1}{l|}{}                     & \multicolumn{1}{l|}{}                       & \multicolumn{1}{l|}{MS-Marco\_place \cite{msmarco} \cite{place-qa}}              & \multicolumn{1}{c|}{\ding{56}}                   & \multicolumn{1}{c|}{\ding{56}}        & \multicolumn{1}{c|}{\ding{56}}            & \multicolumn{1}{r|}{23 513}            & \multicolumn{1}{r|}{4 149}           & 2 907                                 & -      & 6.67   & \multicolumn{1}{c|}{6.9}    & \multicolumn{1}{c|}{\ding{56}}  & \multicolumn{1}{c|}{\ding{56}}  & \multicolumn{1}{c|}{\ding{56}}  & \multicolumn{1}{c|}{\ding{56}}   & \multicolumn{1}{c|}{\ding{56}}                  & \multicolumn{1}{c|}{\ding{56}}                  & \multicolumn{1}{c|}{\ding{56}}               & \multicolumn{1}{c|}{\ding{56}}                   & \multicolumn{1}{c|}{\ding{56}}            & \multicolumn{1}{c|}{\ding{73}}               & \multicolumn{1}{c|}{\ding{51}}                   & \multicolumn{1}{c|}{\ding{56}}           & \multicolumn{1}{c|}{\ding{56}}           & \multicolumn{1}{c|}{\ding{56}}           & \multicolumn{1}{c|}{\ding{56}}           & \multicolumn{1}{c|}{\ding{56}}                 & \multicolumn{1}{c}{\ding{56}}                        \\ \midrule

\multicolumn{1}{l|}{\textbf{Reasoning}}   & \multicolumn{1}{l|}{Complex Scenario QA}    & \multicolumn{1}{l|}{GeoSQA \cite{geosqa}}                       & \multicolumn{1}{c|}{\ding{56}}                   & \multicolumn{1}{c|}{\ding{51}}         & \multicolumn{1}{c|}{\ding{51}}           & \multicolumn{1}{r|}{2 644}                 & \multicolumn{1}{r|}{628}               & 838                                 & 92.51  & 35.93  & \multicolumn{1}{c|}{1.00}   & \multicolumn{1}{c|}{\ding{73}}  & \multicolumn{1}{c|}{\ding{56}}  & \multicolumn{1}{c|}{\ding{56}}  & \multicolumn{1}{c|}{\ding{56}}   & \multicolumn{1}{c|}{\ding{56}}                  & \multicolumn{1}{c|}{\ding{56}}                  & \multicolumn{1}{c|}{\ding{56}}               & \multicolumn{1}{c|}{\ding{56}}                   & \multicolumn{1}{c|}{\ding{56}}            & \multicolumn{1}{c|}{\ding{56}}               & \multicolumn{1}{c|}{\ding{56}}                   & \multicolumn{1}{c|}{\ding{56}}           & \multicolumn{1}{c|}{\ding{56}}           & \multicolumn{1}{c|}{\ding{56}}           & \multicolumn{1}{c|}{\ding{56}}           & \multicolumn{1}{c|}{\ding{56}}                 & \multicolumn{1}{c}{\ding{56}}                        \\
\multicolumn{1}{l|}{}                     & \multicolumn{1}{l|}{}                       & \multicolumn{1}{l|}{GKMC \cite{GKMC}}                         & \multicolumn{1}{c|}{\ding{56}}                   & \multicolumn{1}{c|}{\ding{51}}        & \multicolumn{1}{c|}{\ding{51}}            & \multicolumn{1}{r|}{-}                 & \multicolumn{1}{r|}{-}               & 1 600                                 & 50.87  & 37.67  & \multicolumn{1}{c|}{1.00}   & \multicolumn{1}{c|}{\ding{73}}  & \multicolumn{1}{c|}{\ding{56}}  & \multicolumn{1}{c|}{\ding{56}}  & \multicolumn{1}{c|}{\ding{56}}   & \multicolumn{1}{c|}{\ding{56}}                  & \multicolumn{1}{c|}{\ding{56}}                  & \multicolumn{1}{c|}{\ding{56}}               & \multicolumn{1}{c|}{\ding{56}}                   & \multicolumn{1}{c|}{\ding{56}}            & \multicolumn{1}{c|}{\ding{56}}               & \multicolumn{1}{c|}{\ding{56}}                   & \multicolumn{1}{c|}{\ding{56}}           & \multicolumn{1}{c|}{\ding{56}}           & \multicolumn{1}{c|}{\ding{56}}           & \multicolumn{1}{c|}{\ding{56}}           & \multicolumn{1}{c|}{\ding{56}}                 & \multicolumn{1}{c}{\ding{56}}                        \\ \cline{2-29}

\multicolumn{1}{l|}{}                     & \multicolumn{1}{l|}{Spatial Reasoning}      & \multicolumn{1}{l|}{SpatialEvalLLM \cite{spatialEvalLLM}}               & \multicolumn{1}{c|}{\ding{51}}                   & \multicolumn{1}{c|}{\ding{51}}        & \multicolumn{1}{c|}{\ding{56}}            & \multicolumn{1}{r|}{-}                 & \multicolumn{1}{r|}{-}               & 1 400                                 & 138.01 & 4.50   & \multicolumn{1}{c|}{1.59}   & \multicolumn{1}{c|}{\ding{73}}  & \multicolumn{1}{c|}{\ding{56}}  & \multicolumn{1}{c|}{\ding{56}}  & \multicolumn{1}{c|}{\ding{56}}   & \multicolumn{1}{c|}{\ding{56}}                  & \multicolumn{1}{c|}{\ding{56}}                  & \multicolumn{1}{c|}{\ding{56}}               & \multicolumn{1}{c|}{\ding{56}}                   & \multicolumn{1}{c|}{\ding{56}}            & \multicolumn{1}{c|}{\ding{56}}               & \multicolumn{1}{c|}{\ding{56}}                   & \multicolumn{1}{c|}{\ding{56}}           & \multicolumn{1}{c|}{\ding{56}}           & \multicolumn{1}{c|}{\ding{56}}           & \multicolumn{1}{c|}{\ding{56}}           & \multicolumn{1}{c|}{\ding{56}}                 & \multicolumn{1}{c}{\ding{56}}                        \\
\multicolumn{1}{l|}{}                     & \multicolumn{1}{l|}{}                       & \multicolumn{1}{l|}{SpartUN \cite{spartun}}                      & \multicolumn{1}{c|}{\ding{51}}                   & \multicolumn{1}{c|}{\ding{51}}            & \multicolumn{1}{c|}{\ding{56}}        & \multicolumn{1}{r|}{37 095}            & \multicolumn{1}{r|}{5 600}           & 5 551                                 & 85.88  & 10.20  & \multicolumn{1}{c|}{1.29}   & \multicolumn{1}{c|}{\ding{73}}  & \multicolumn{1}{c|}{\ding{56}}  & \multicolumn{1}{c|}{\ding{56}}  & \multicolumn{1}{c|}{\ding{56}}   & \multicolumn{1}{c|}{\ding{56}}                  & \multicolumn{1}{c|}{\ding{56}}                  & \multicolumn{1}{c|}{\ding{56}}               & \multicolumn{1}{c|}{\ding{56}}                   & \multicolumn{1}{c|}{\ding{56}}            & \multicolumn{1}{c|}{\ding{56}}               & \multicolumn{1}{c|}{\ding{56}}                   & \multicolumn{1}{c|}{\ding{56}}           & \multicolumn{1}{c|}{\ding{56}}           & \multicolumn{1}{c|}{\ding{56}}           & \multicolumn{1}{c|}{\ding{56}}           & \multicolumn{1}{c|}{\ding{56}}                 & \multicolumn{1}{c}{\ding{56}}                        \\
\multicolumn{1}{l|}{}                     & \multicolumn{1}{l|}{}                       & \multicolumn{1}{l|}{StepGame \cite{stepgame}}                     & \multicolumn{1}{c|}{\ding{51}}                   & \multicolumn{1}{c|}{\ding{51}}       & \multicolumn{1}{c|}{\ding{56}}             & \multicolumn{1}{r|}{50 000}            & \multicolumn{1}{r|}{5 000}           & 100 000                               & 83.35  & 12.00  & \multicolumn{1}{c|}{1.00}   & \multicolumn{1}{c|}{\ding{73}}  & \multicolumn{1}{c|}{\ding{56}}  & \multicolumn{1}{c|}{\ding{56}}  & \multicolumn{1}{c|}{\ding{56}}   & \multicolumn{1}{c|}{\ding{56}}                  & \multicolumn{1}{c|}{\ding{56}}                  & \multicolumn{1}{c|}{\ding{56}}               & \multicolumn{1}{c|}{\ding{56}}                   & \multicolumn{1}{c|}{\ding{56}}            & \multicolumn{1}{c|}{\ding{56}}               & \multicolumn{1}{c|}{\ding{56}}                   & \multicolumn{1}{c|}{\ding{56}}           & \multicolumn{1}{c|}{\ding{56}}           & \multicolumn{1}{c|}{\ding{56}}           & \multicolumn{1}{c|}{\ding{56}}           & \multicolumn{1}{c|}{\ding{56}}                 & \multicolumn{1}{c}{\ding{56}}                        \\ \midrule

\multicolumn{1}{l|}{\textbf{Application}} & \multicolumn{1}{l|}{POI Recommendation}     & \multicolumn{1}{l|}{TourismQA \cite{tourismQA} \cite{locationawaremodularbiencoder}}                    & \multicolumn{1}{c|}{\ding{56}}                   & \multicolumn{1}{c|}{\ding{56}}       & \multicolumn{1}{c|}{\ding{56}}             & \multicolumn{1}{r|}{19 762}            & \multicolumn{1}{r|}{2 109}           & 2 153                               & -      & 75.69  & \multicolumn{1}{c|}{4.09}   & \multicolumn{1}{c|}{\ding{56}}  & \multicolumn{1}{c|}{\ding{56}}  & \multicolumn{1}{c|}{\ding{56}}  & \multicolumn{1}{c|}{\ding{56}}   & \multicolumn{1}{c|}{\ding{56}}                  & \multicolumn{1}{c|}{\ding{51}}                  & \multicolumn{1}{c|}{\ding{51}}               & \multicolumn{1}{c|}{\ding{51}}                   & \multicolumn{1}{c|}{\ding{51}}            & \multicolumn{1}{c|}{\ding{73}}               & \multicolumn{1}{c|}{\ding{51}}                   & \multicolumn{1}{c|}{\ding{56}}           & \multicolumn{1}{c|}{\ding{56}}           & \multicolumn{1}{c|}{\ding{56}}           & \multicolumn{1}{c|}{\ding{56}}           & \multicolumn{1}{c|}{\ding{56}}                 & \multicolumn{1}{c}{\ding{56}}                        \\
\multicolumn{1}{l|}{}                     & \multicolumn{1}{l|}{}                       & \multicolumn{1}{l|}{NY-POI \cite{NY} \cite{GETNext} \cite{NY-preprocessed} \cite{llm-poi}}                       & \multicolumn{1}{c|}{\ding{56}}                   & \multicolumn{1}{c|}{\ding{51}}       & \multicolumn{1}{c|}{\ding{51}}             & \multicolumn{1}{r|}{-}                 & \multicolumn{1}{r|}{-}               & 1 347                                 & 136.13 & 396.28 & \multicolumn{1}{c|}{1.0}    & \multicolumn{1}{c|}{\ding{73}}  & \multicolumn{1}{c|}{\ding{51}}  & \multicolumn{1}{c|}{\ding{51}}  & \multicolumn{1}{c|}{\ding{51}}   & \multicolumn{1}{c|}{\ding{56}}                  & \multicolumn{1}{c|}{\ding{56}}                  & \multicolumn{1}{c|}{\ding{56}}               & \multicolumn{1}{c|}{\ding{56}}                   & \multicolumn{1}{c|}{\ding{56}}            & \multicolumn{1}{c|}{\ding{56}}               & \multicolumn{1}{c|}{\ding{56}}                   & \multicolumn{1}{c|}{\ding{56}}           & \multicolumn{1}{c|}{\ding{56}}           & \multicolumn{1}{c|}{\ding{56}}           & \multicolumn{1}{c|}{\ding{56}}           & \multicolumn{1}{c|}{\ding{56}}                 & \multicolumn{1}{c}{\ding{56}}                        \\ \cline{2-29} 
\multicolumn{1}{l|}{}                     & \multicolumn{1}{l|}{Pathfinding}           & \multicolumn{1}{l|}{GridRoute \cite{gridroute}}                    & \multicolumn{1}{c|}{\ding{51}}                   & \multicolumn{1}{c|}{\ding{51}}       & \multicolumn{1}{c|}{\ding{56}}             & \multicolumn{1}{r|}{-}                 & \multicolumn{1}{r|}{-}               & 300                                   & 112.67 & -      & \multicolumn{1}{c|}{29.83}  & \multicolumn{1}{c|}{\ding{56}}  & \multicolumn{1}{c|}{\ding{56}}  & \multicolumn{1}{c|}{\ding{56}}  & \multicolumn{1}{c|}{\ding{56}}   & \multicolumn{1}{c|}{\ding{56}}                  & \multicolumn{1}{c|}{\ding{56}}                  & \multicolumn{1}{c|}{\ding{56}}               & \multicolumn{1}{c|}{\ding{56}}                   & \multicolumn{1}{c|}{\ding{56}}            & \multicolumn{1}{c|}{\ding{56}}               & \multicolumn{1}{c|}{\ding{56}}                   & \multicolumn{1}{c|}{\ding{51}}           & \multicolumn{1}{c|}{\ding{51}}           & \multicolumn{1}{c|}{\ding{73}}           & \multicolumn{1}{c|}{\ding{51}}           & \multicolumn{1}{c|}{\ding{51}}                 & \multicolumn{1}{c}{\ding{51}}                        \\
\multicolumn{1}{l|}{}                     & \multicolumn{1}{l|}{}                       & \multicolumn{1}{l|}{PPNL\_single \cite{ppnl}}                 & \multicolumn{1}{c|}{\ding{51}}                   & \multicolumn{1}{c|}{\ding{51}}          & \multicolumn{1}{c|}{\ding{56}}         & \multicolumn{1}{r|}{16 032}            & \multicolumn{1}{r|}{2 004}           & 19 044                                & 9.86   & -      & \multicolumn{1}{c|}{10.21}  & \multicolumn{1}{c|}{\ding{56}}  & \multicolumn{1}{c|}{\ding{56}}  & \multicolumn{1}{c|}{\ding{56}}  & \multicolumn{1}{c|}{\ding{56}}   & \multicolumn{1}{c|}{\ding{56}}                  & \multicolumn{1}{c|}{\ding{56}}                  & \multicolumn{1}{c|}{\ding{56}}               & \multicolumn{1}{c|}{\ding{56}}                   & \multicolumn{1}{c|}{\ding{56}}            & \multicolumn{1}{c|}{\ding{56}}               & \multicolumn{1}{c|}{\ding{56}}                   & \multicolumn{1}{c|}{\ding{51}}           & \multicolumn{1}{c|}{\ding{51}}           & \multicolumn{1}{c|}{\ding{73}}           & \multicolumn{1}{c|}{\ding{51}}           & \multicolumn{1}{c|}{\ding{51}}                 & \multicolumn{1}{c}{\ding{51}}                        \\
\multicolumn{1}{l|}{}                     & \multicolumn{1}{l|}{}                       & \multicolumn{1}{l|}{PPNL\_multi \cite{ppnl}}                  & \multicolumn{1}{c|}{\ding{51}}                   & \multicolumn{1}{c|}{\ding{51}}         & \multicolumn{1}{c|}{\ding{56}}          & \multicolumn{1}{r|}{53 440}            & \multicolumn{1}{r|}{6 680}           & 55 440                                & 14.99  & -      & \multicolumn{1}{c|}{33.57}  & \multicolumn{1}{c|}{\ding{56}}  & \multicolumn{1}{c|}{\ding{56}}  & \multicolumn{1}{c|}{\ding{56}}  & \multicolumn{1}{c|}{\ding{56}}   & \multicolumn{1}{c|}{\ding{56}}                  & \multicolumn{1}{c|}{\ding{56}}                  & \multicolumn{1}{c|}{\ding{56}}               & \multicolumn{1}{c|}{\ding{56}}                   & \multicolumn{1}{c|}{\ding{56}}            & \multicolumn{1}{c|}{\ding{56}}               & \multicolumn{1}{c|}{\ding{56}}                   & \multicolumn{1}{c|}{\ding{51}}           & \multicolumn{1}{c|}{\ding{51}}           & \multicolumn{1}{c|}{\ding{73}}           & \multicolumn{1}{c|}{\ding{51}}           & \multicolumn{1}{c|}{\ding{51}}                 & \multicolumn{1}{c}{\ding{51}}                        \\ \hline

\multicolumn{1}{l|}{\textbf{Total}}    & \multicolumn{1}{c|}{-}      & \multicolumn{1}{c|}{-}                      & \multicolumn{1}{c|}{-}                            & \multicolumn{1}{c|}{-}                   & \multicolumn{1}{c|}{-}                  & \multicolumn{1}{r|}{203 014}           & \multicolumn{1}{r|}{26 220}          & 191 807                              & \multicolumn{3}{c}{421 041}                   &                         &                         &                         &                          &                                         &                                         &                                      &                                          &                                   &                                      &                                          &                                  &                                  &                                  &                                  &                                        & \multicolumn{1}{c}{}                         \\ \bottomrule
\end{tabular}%
}
\caption{\benchName\ description (\ding{73} points to the main metric used to evaluate the dataset). Cognitive level (Cogn. lvl), Synthetic (Syn.), Scenario (Sce.), Transformation (Trs.), Coordinates Accuracy (C Acc), Precision (Prec), Recall (Rec), Precision-Recall Mean (P-R Mean), BERT-Score (B-Sco), Compliance Ratio (C Rat), Success Ratio (S Ratio), Optimal Ratio (O Ratio), Feasible Ratio (F Rat), Distance (Dist) and Unreachable Accuracy (U Acc) }
\label{tab:description}
\vspace{-.8cm}
\end{table*}

\subsection{Data Collection}

We collected twelve publicly available datasets directly from author-provided links. \textit{TourismQA} \cite{tourismQA} could not be regenerated from its original code and was therefore retrieved from a work using it \cite{locationawaremodularbiencoder}. Where necessary, we applied transformations — including those from prior works — to make datasets suitable for LLM evaluation, and partitioned some into subdatasets to prevent overlap across tasks. Transformation details are given alongside each dataset's description in the next section.

We adopt a taxonomy inspired by Xu et al. \cite{evaluatingllm}, organizing our benchmark into three cognitive levels — \textit{Knowledge}, \textit{Reasoning}, and \textit{Application} — to clearly divide tasks according to the skills required to complete them. \textit{Knowledge} tasks comprise factoid questions answerable by querying a geographical database, split into four subtasks by answer type: coordinates, real numbers, Yes/No, or place names. \textit{Reasoning} tasks require the model to apply those concepts: \textit{Complex Scenario QA} involves processing knowledge and facts in an MCQ format, while \textit{Spatial Reasoning} requires understanding distance, topology, and direction. \textit{Application} tasks build further on these skills for real-world use cases: \textit{POI Recommendation} combines world knowledge, spatial reasoning, and user preference processing, while \textit{Pathfinding} additionally incorporates a temporal factor to construct paths between points.

\subsection{Tasks and datasets}

We now describe, for each cognitive level, the datasets and tasks that belong to it.

\subsubsection{Knowledge datasets}

\begin{enumerate}[label=\textbf{\alph*)}, itemsep=0pt, topsep=0pt, leftmargin=10pt]
    \item \textbf{GeoQuestions1089} \cite{geoquestions} consists of 1089 geographical factoid questions covering four regions: the United States, the United Kingdom, Ireland, and Greece, with varying degrees of complexity. Originally designed for natural language to SPARQL translation over the YAGO2 and YAGO2Geo knowledge bases,\footnote{YAGO2 is a multi-domain knowledge base focusing on the spatio-temporal aspect of data. YAGO2Geo is an extension of the latter, providing new and more accurate geographical information.} the dataset also contains the raw query responses, which we processed and cleaned to extract only the useful data.
    \item \textbf{GeoQuery} \cite{geoquery} was originally developed to assess natural language to database query parsing. It comprises simple questions about geographical facts concerning the United States, drawn from the \textit{GeoBase} knowledge base (1996). We use the version made available by Finegan-Dollak et al. \cite{texttosql} for text-to-SQL evaluation.
    \item \textbf{Ms-Marco} \cite{msmarco} is a large-scale dataset of 1,000,000 question-answer pairs collected from Bing. We use a geographical subset of the QA task (version 2.1), retaining only LOCATION-category questions that mention at least one real geographical entity, following Hamzei et al. \cite{place-qa}.
\end{enumerate}

\subsubsection{Reasoning datasets}

\begin{enumerate}[label=\textbf{\alph*)}, itemsep=0pt, topsep=0pt, leftmargin=10pt]
    \item \textbf{GeoSQA} \cite{geosqa} consists of 4,110 multiple-choice questions from Gaokao geography examinations,\footnote{Chinese equivalent of the Baccalaureate or High School Diploma} each accompanied by a scenario and a diagram annotated in text by the authors. Originally in Chinese, we translated the dataset into English using \textit{Google Translate}.
    \item \textbf{GKMC} \cite{GKMC} is also a scenario-based multiple-choice dataset drawn from the Gaokao, but without diagrams, distinguishing it from GeoSQA. It was similarly translated from Chinese into English.
    \item \textbf{SpatialEvalLLM} \cite{spatialEvalLLM} evaluates spatial understanding by placing the LLM within a shaped grid of objects and asking it to identify the object at the end of a described path.
    \item \textbf{SpartUN} \cite{spartun} evaluates spatial reasoning by constructing scenarios of objects connected topologically or directionally, then asking either \textbf{Boolean} (truthfulness of a relationship) or \textbf{Relational} (choice of best-fitting relationship) questions.
    \item \textbf{StepGame} \cite{stepgame} is a synthetic dataset where the model must infer the directional relationship between two points from a scenario describing intermediate placements. Questions are categorised by the number of reasoning hops required, with increasing difficulty.
\end{enumerate}

\begin{table}[t!]
\resizebox{0.48\textwidth}{!}{%
\begin{tabular}{llcc|cccccccc}
\multicolumn{1}{l|}{\multirow{2}{*}{\textbf{Cog. lvl}}} & \multicolumn{1}{l|}{\multirow{2}{*}{\textbf{Dataset}}} & \multicolumn{1}{c|}{\multirow{2}{*}{\textbf{\ding{73} Metric}}} & \multirow{2}{*}{\rotatebox{90}{\textbf{Dir.}}} & \multicolumn{3}{c}{\textbf{Qwen3 (w\textbackslash{}o thinking)}}       & \multicolumn{3}{c}{\textbf{Qwen3 (with thinking)}}             & \multicolumn{2}{c}{\textbf{GPT-OSS}} \\ \cline{5-12} 
\multicolumn{1}{l|}{}                                     & \multicolumn{1}{l|}{}                                  & \multicolumn{1}{c|}{}                                     &                                & 0.6B           & 1.7B           & 8B                                  & 0.6B           & 1.7B  & 8B                                   & 20B               & 120B             \\ \hline
\multicolumn{1}{l|}{\textbf{Know.}}           & \multicolumn{1}{l|}{GeoQ.1089\_coord}           & \multicolumn{1}{c|}{C Acc}                                & ↑                              & 0.01           & 0.14           & \multicolumn{1}{c|}{0.39}           & 0.02           & 0.11  & \multicolumn{1}{c|}{0.43}            & \underline{0.48}    & \textbf{0.72}    \\ \cline{2-12} 
\multicolumn{1}{l|}{}                                     & \multicolumn{1}{l|}{GeoQ.1089\_YN}              & \multicolumn{1}{c|}{Acc}                                  & ↑                              & 0.51           & 0.55           & \multicolumn{1}{c|}{0.64}           & 0.55           & 0.63  & \multicolumn{1}{c|}{0.67}            & \textbf{0.77}     & \underline{0.73}   \\ \cline{2-12} 
\multicolumn{1}{l|}{}                                     & \multicolumn{1}{l|}{GeoQ.1089\_regress.}      & \multicolumn{1}{c|}{Median}                               & ↓                              & 1.1k           & 0.5k           & \multicolumn{1}{c|}{1.2k}           & 0.2k           & 0.5k  & \multicolumn{1}{c|}{0.4k}            & \underline{0.1k}    & \textbf{0.1k}    \\
\multicolumn{1}{l|}{}                                     & \multicolumn{1}{l|}{GeoQuery\_regress.}              & \multicolumn{1}{c|}{Median}                               & ↓                              & 17.4k          & 12.2k          & \multicolumn{1}{c|}{\textbf{5.7k}}  & 11.3k          & 8.6k  & \multicolumn{1}{c|}{14.1k}           & 9.3k              & \underline{6.1k}   \\ \cline{2-12} 
\multicolumn{1}{l|}{}                                     & \multicolumn{1}{l|}{GeoQ.1089\_place}           & \multicolumn{1}{c|}{Median}                               & ↓                              & 33.50          & 40.50          & \multicolumn{1}{c|}{37.00}          & \textbf{26.50} & 28.50 & \multicolumn{1}{c|}{\underline{28.00}} & 29.50             & 29.50            \\
\multicolumn{1}{l|}{}                                     & \multicolumn{1}{l|}{GeoQuery\_place}                   & \multicolumn{1}{c|}{Median}                               & ↓                              & 12.50          & 14.50          & \multicolumn{1}{c|}{8.00}           & 10.00          & 9.00  & \multicolumn{1}{c|}{\underline{6.25}}  & 6.00              & \textbf{5.00}    \\
\multicolumn{1}{l|}{}                                     & \multicolumn{1}{l|}{MS-Marco\_place}                   & \multicolumn{1}{c|}{Bleu-1}                               & ↑                              & \underline{0.18} & 0.06           & \multicolumn{1}{c|}{0.11}           & \textbf{0.19}  & 0.08  & \multicolumn{1}{c|}{0.06}            & 0.08              & 0.07             \\ \hline
\multicolumn{1}{l|}{\textbf{Reas.}}         & \multicolumn{1}{l|}{GeoSQA}                            & \multicolumn{1}{c|}{Acc}                                  & ↑                              & 0.21           & 0.38           & \multicolumn{1}{c|}{0.55}           & 0.37           & 0.50  & \multicolumn{1}{c|}{\textbf{0.63}}   & 0.53              & \underline{0.57}   \\
\multicolumn{1}{l|}{}                                     & \multicolumn{1}{l|}{GKMC}                              & \multicolumn{1}{c|}{Acc}                                  & ↑                              & 0.28           & 0.59           & \multicolumn{1}{c|}{0.77}           & 0.50           & 0.59  & \multicolumn{1}{c|}{\textbf{0.82}}   & 0.73              & \underline{0.79}   \\ \cline{2-12} 
\multicolumn{1}{l|}{}                                     & \multicolumn{1}{l|}{SpatialEvalLLM}                    & \multicolumn{1}{c|}{Acc}                                  & ↑                              & 0.02           & 0.03           & \multicolumn{1}{c|}{0.04}           & 0.03           & 0.15  & \multicolumn{1}{c|}{0.29}            & \underline{0.31}    & \textbf{0.37}    \\
\multicolumn{1}{l|}{}                                     & \multicolumn{1}{l|}{SpartUN}                           & \multicolumn{1}{c|}{Acc}                                  & ↑                              & 0.32           & 0.32           & \multicolumn{1}{c|}{0.46}           & 0.38           & 0.46  & \multicolumn{1}{c|}{0.64}            & \underline{0.77}    & \textbf{0.86}    \\
\multicolumn{1}{l|}{}                                     & \multicolumn{1}{l|}{StepGame}                          & \multicolumn{1}{c|}{Acc}                                  & ↑                              & 0.15           & 0.15           & \multicolumn{1}{c|}{0.23}           & 0.22           & 0.36  & \multicolumn{1}{c|}{0.53}            & \textbf{0.72}     & \underline{0.64}   \\ \hline
\multicolumn{1}{l|}{\textbf{Appl.}}   & \multicolumn{1}{l|}{TourismQA}                         & \multicolumn{1}{c|}{Bleu-1}                               & ↑                              & 0.03           & 0.06           & \multicolumn{1}{c|}{\underline{0.07}} & 0.04           & 0.01  & \multicolumn{1}{c|}{0.01}            & \underline{0.07}    & \textbf{0.08}    \\
\multicolumn{1}{l|}{}                                     & \multicolumn{1}{l|}{NY-POI}                            & \multicolumn{1}{c|}{Acc}                                  & ↑                              & 0.32           & \underline{0.39} & \multicolumn{1}{c|}{0.33}           & 0.33           & 0.33  & \multicolumn{1}{c|}{\underline{0.39}}  & \textbf{0.40}     & \underline{0.39}   \\ \cline{2-12} 
\multicolumn{1}{l|}{}                                     & \multicolumn{1}{l|}{GridRoute}                         & \multicolumn{1}{c|}{O Rat}                                & ↑                              & 0.13           & 0.31           & \multicolumn{1}{c|}{0.51}           & 0.41           & 0.66  & \multicolumn{1}{c|}{\textbf{0.81}}   & 0.67              & \underline{0.80}   \\
\multicolumn{1}{l|}{}                                     & \multicolumn{1}{l|}{PPNL\_single}                      & \multicolumn{1}{c|}{O Rat}                                & ↑                              & 0.03           & 0.39           & \multicolumn{1}{c|}{0.53}           & 0.60           & 0.83  & \multicolumn{1}{c|}{\textbf{0.92}}   & 0.83              & \underline{0.90}   \\
\multicolumn{1}{l|}{}                                     & \multicolumn{1}{l|}{PPNL\_multi}                       & \multicolumn{1}{c|}{O Rat}                                & ↑                              & 0.00           & 0.02           & \multicolumn{1}{c|}{0.04}           & 0.07           & 0.23  & \multicolumn{1}{c|}{\textbf{0.62}}   & 0.41              & \underline{0.57}   \\ \hline
\multicolumn{4}{c|}{\textbf{Total wins}}                                                                                                                                                                                 & 0              & 0              & \multicolumn{1}{c|}{1}              & 2              & 0     & \multicolumn{1}{c|}{5}               & 3                 & 6                \\ \hline
\multicolumn{4}{c|}{\textbf{Total runner-up}}                                                                                                                                                                            & 1              & 1              & \multicolumn{1}{c|}{1}              & 0              & 0     & \multicolumn{1}{c|}{3}               & 5                 & 9                \\ \hline
\end{tabular}%
}
\caption{Main results on \benchName. For each subdataset, we only show the main metric specified in Table \ref{tab:description}. Cognitive level (Cog. lvl), Knowledge (Know.), Reasoning (Reas.), Application (Appl.), Direction (Dir.).\label{tab:resultsmain}}
\vspace{-.8cm}
\end{table}

\subsubsection{Application datasets}

\begin{enumerate}[label=\textbf{\alph*)}, itemsep=0pt, topsep=0pt, leftmargin=10pt]
    \item \textbf{TourismQA} \cite{tourismQA} is a POI recommendation dataset built from tourist site reviews across fifty cities worldwide. Given a tourist's question and available reviews, the task is to predict relevant POIs. We retrieved it from the repository of Li et al. \cite{locationawaremodularbiencoder} as the original code could not be used to regenerate it.
    \item \textbf{NY-POI} \cite{NY} is derived from Foursquare check-ins and contains geolocated POIs with user--POI interaction sequences and timestamps. The task is to predict the next POI in a trajectory given the user's habits and visit history. We follow the task definition and pre-processing pipeline of \cite{NY}, \cite{GETNext}, and \cite{NY-preprocessed}.
    \item \textbf{GridRoute} \cite{gridroute} assesses pathfinding by asking the model to return a valid sequence of adjacent grid coordinates from a starting point A to an end point B, avoiding obstacles and without diagonal movements.
    \item \textbf{PPNL} \cite{ppnl} extends the pathfinding task to a multi-objective setting, where the model must pass through an unordered list of intermediate points to produce a correct solution, making it more difficult than GridRoute.
\end{enumerate}

\section{Experiments and results}

\label{metrics}

\subsection{Baselines}
We explored the use of basic LLMs from the Qwen3 family with various sizes (Qwen3-0.6B, Qwen3-1.7B, Qwen3-8B), with and without thinking. Additionally, we also explored two larger models in the GPT-OSS-20B and 120B versions.

\subsection{Metrics}

To avoid restricting evaluation to MCQ and Yes/No questions, which introduce a bias in the assessment of model abilities. However, the evaluation of text generation can be complex as they are open questions.
For example, the subdatasets GeoQuestions1089\_regression and Geo-
Query\_regression expect as an answer a list of real numbers. For
such cases, we introduce custom metrics for open-ended tasks.\footnote{All metrics are available in a \href{https://huggingface.co/collections/rfr2003/geobenchllm-metrics}{Hugging Face Collection} and can be used with the \textit{evaluate} library: \url{https://huggingface.co/collections/rfr2003/geobenchllm-metrics}}

\begin{itemize}[itemsep=0pt, topsep=0pt, leftmargin=10pt]
    \item \textbf{Coordinates Accuracy}: used for \textit{Coordinates prediction}, a predicted coordinate $p$ is correct if it falls within a circle of radius $r$ centred at the gold coordinate $g$:
    \begin{equation*}
        Coord\_Acc =
    \begin{cases}
    1 & \text{if } d(p, g) \leq r, \\[4pt]
    0 & \text{otherwise}
    \end{cases}
    \end{equation*}
    where $d$ denotes the haversine distance, and $r$ the tolerance radius.

    \item \textbf{Precision, Recall, P-R Mean and Median}: used for \textit{Place prediction} (except \textit{Ms-Marco\_place}) and \textit{Regression}, where both the reference and predicted answers are lists of values. \textit{Precision} measures how close predictions are to references; \textit{Recall} measures coverage of references; \textit{P-R Mean} is their average. The median of P-R Mean across questions is our main metric, chosen to reduce the impact of outliers:
    \begin{equation*}
        D_{i,j} = d(g_i, p_j)
    \quad \text{for } 1 \le i \le |G|,\ 1 \le j \le |P|
    \end{equation*}
    \begin{equation*}
        \text{Prec} = \sum_{j=1}^{|P|} \min_{1 \le i \le |G|} D_{i,j}
        \qquad
        \text{Rec} = \sum_{i=1}^{|G|} \min_{1 \le j \le |P|} D_{i,j}
    \end{equation*}
    where $D$ is the distance matrix between reference values $G$ and predicted values $P$.

    \item \textbf{Compliance Ratio}: measures how often the model's output follows the required format (list of grid coordinates) in the \textit{Pathfinding} task.

    \item \textbf{Feasible, Success and Optimal Ratios} (from Aghzal et al.~\cite{ppnl}): a path is \textit{feasible} if it stays within grid boundaries and avoids obstacles, \textit{successful} if it also reaches the goal, and \textit{optimal} if it does so in the minimum number of moves. Each ratio is computed over reachable paths. \textit{Optimal Ratio} is our main metric as the most discriminating.

    \item \textbf{Unreachable Accuracy}: measures how often the model correctly detects that a goal is unreachable.

    \item \textbf{Distance}: the minimum number of additional moves needed to turn a feasible path into a successful one.

    \item \textbf{Accuracy}: used for closed-answer tasks (\textit{Yes/No questions}, \textit{Complex Scenario QA}, \textit{Spatial Reasoning}, and \textit{NY-POI}).

    \item \textbf{Bleu-1 and Bert-Score}: used for \textit{Ms-Marco\_place} and \textit{TourismQA}. Bleu-1 is our main metric as it is more discriminating than Bert-Score.

\end{itemize}

\begin{figure}
    \includegraphics[width=\columnwidth]{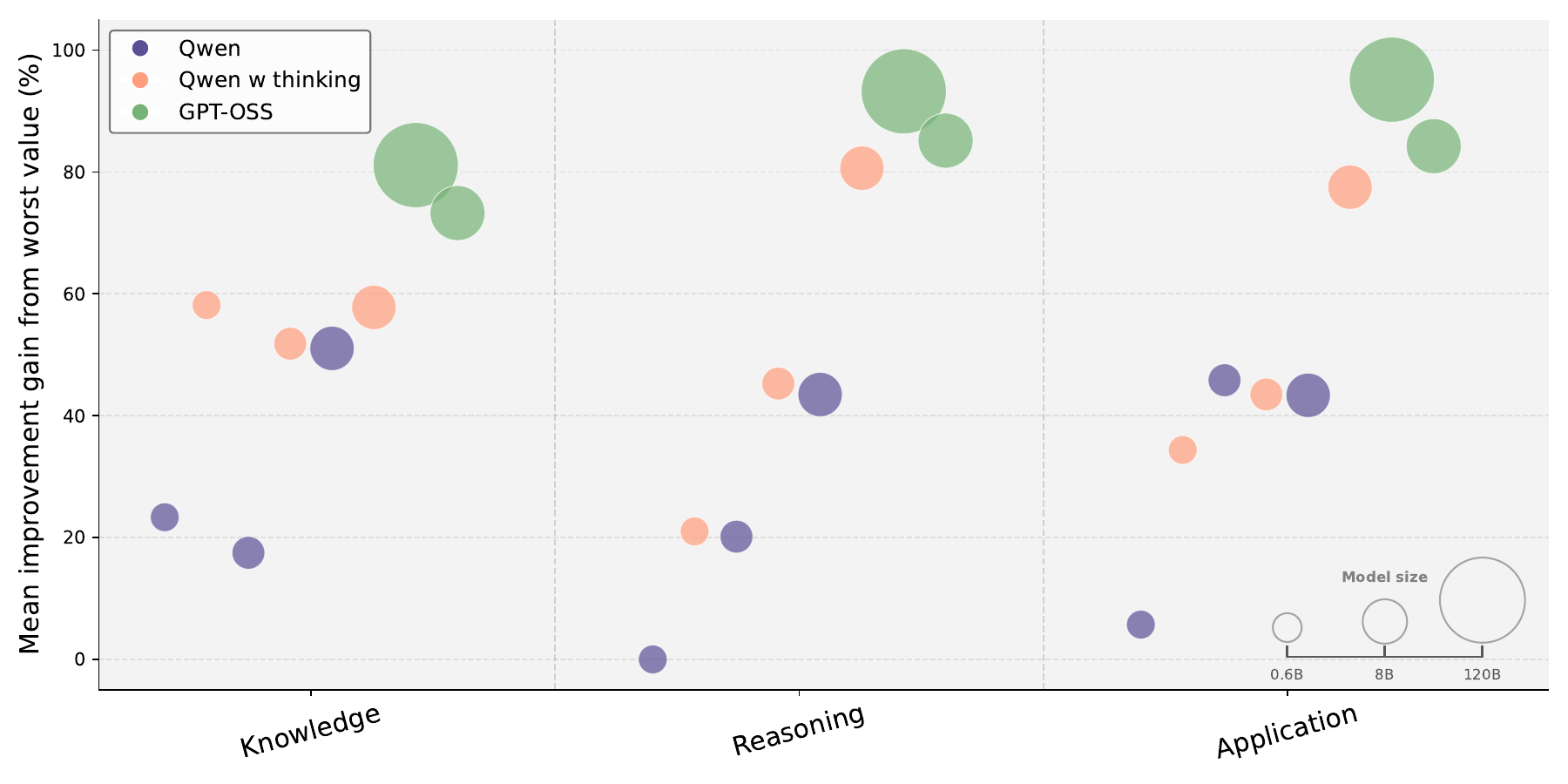}
    \caption{Mean improvement gain for each model on each cognitive level. For each subdataset, the improvement gain from the worst result was computed. We plot the mean over these gains for subdatasets of the same cognitive level.}
    \label{fig:code}
    \vspace{-0.6cm}
\end{figure}

\subsection{Results}
Table \ref{tab:resultsmain} presents our main results.\footnote{A notebook including the local calculation of the results for the GeoSQA dataset is available in our \href{https://github.com/Rfr2003/GeoBenchLLM/blob/main/example.ipynb}{GitHub}.} Note that we only allowed low thinking budgets to GPT-OSS-20B and 120B. Unsurprisingly, these larger models still manage to achieve the best results in nine of the seventeen subdatasets. However, a relatively small model, Qwen3-8B, is capable of achieving similar performance when thinking mode is activated. It even surpasses the largest model on five of the ten subdatasets belonging to the \textit{Reasoning} and \textit{Application} cognitive levels, achieving 0.62 in accuracy on \textit{PPNL\_multi} while the largest model only attains 0.57, for example. This suggests that for tasks where reasoning plays a central role, thinking is more important than size. On the other hand, GPT-OSS models lead in four out of the seven subdatasets of the \textit{Knowledge} level, often by a wide margin. On \textit{GeoQuestions1089\_coord}, GPT-OSS-120B outperforms the best Qwen model by 0.29 in coordinates accuracy. We interpret this as indicating that, without external tools and databases, the size of the model plays a major role in embedding real-world knowledge, which is required to answer questions at the \textit{Knowledge} cognitive level. These tendencies are shown in Figure \ref{fig:code}. We observe that for the \textit{Reasoning} and \textit{Application} cognitive levels, the gap between the largest models and Qwen3-8B in thinking mode is very narrow compared to the one at the \textit{Knowledge} level. This gap goes from around 24\% for the \textit{Knowledge} level to 13\% and 18\% for the \textit{Reasoning} and \textit{Application} levels respectively. We can also see the power of reasoning, as the thinking version of the same model almost always surpasses the non-thinking one, sometimes reaching the same performance as the next model in size. It would therefore not be surprising that using a larger model with thinking would yield a marked improvement in performance across all subdatasets.

\section{Conclusion}
This paper presents \benchName, a general benchmark containing twelve datasets for assessing the abilities of large language models in geo-related tasks at different cognitive levels. Our benchmark contains eight tasks ranging from \textit{Coordinates Prediction} to \textit{Pathfinding}. We made sure to make the benchmark as accessible as possible to future research. We also introduced new metrics to evaluate open questions in a generation context. Finally, we present baselines consisting of small models from the Qwen3 family with and without thinking, as well as larger models ranging from 20B to 120B parameters. We found that the parameter gap could be closed using thinking, especially for \textit{Reasoning} and \textit{Application} tasks.

\bibliographystyle{ACM-Reference-Format}
\bibliography{sample-base}

\end{document}